# High-Order Associative Learning Based on Memristive Circuits for Efficient Learning

Shengbo Wang[1], Xuemeng Li[1], Jialin Ding[1], Weihao Ma[1], Ying Wang[2], Luigi Occhipinti[3], Arokia Nathan[3] and Shuo Gao[1]
[1]School of Instrumentation and Optoelectronic Engineering, Beihang University, Beijing, China
[2]School of Biological Science and Medical Engineering, Beihang University, Beijing, China
[3]Department of Engineering, Univeristy of Cambridge, Cambridge, UK
shuo_gao@buaa.edu.cn

*Abstract*—Memristive associative learning has gained significant attention for its ability to mimic fundamental biological learning mechanisms while maintaining system simplicity. In this work, we introduce a high-order memristive associative learning framework with a biologically realistic structure. By utilizing memristors as synaptic modules and their state information to bridge different orders of associative learning, our design effectively establishes associations between multiple stimuli and replicates the transient nature of high-order associative learning. In Pavlov's classical conditioning experiments, our design achieves a 230% improvement in learning efficiency compared to previous works, with memristor power consumption in the synaptic modules remaining below 11 $\mu$W. In large-scale image recognition tasks, we utilize a 20×20 memristor array to represent images, enabling the system to recognize and label test images with semantic information at 100% accuracy. This scalability across different tasks highlights the framework's potential for a wide range of applications, offering enhanced learning efficiency for current memristor-based neuromorphic systems.

*Keywords—neuromorphic systems, memristor, associative learning, bio-inspired computing*

## I. INTRODUCTION

Memristors, with synapse-like characteristics, have been widely employed to mimic essential biological cognitive mechanisms, thereby enhancing the efficiency of current artificial intelligence systems in both energy consumption and learning performance [1]-[5]. Among the various learning mechanisms, associative learning, which allows biological systems to link various sensory inputs and identify complex patterns in the information, has gained significant attention due to its importance in biological cognition systems [6]-[8].

Recently advances have demonstrated the potential of memristors to build high-order associative learning circuits to link multiple stimuli, such as those seen in second-order conditioning in Pavlov's experiments [9]-[12]. However, these methods have yet to replicate the superior learning efficiency of high-order associative learning compared to lower-order learning, i.e., the fewer training rounds to link different stimuli. This limitation results in longer convergence times for memristive systems during learning tasks, reducing their ability to process complex patterns as efficiently as biological systems [13], [7], [14]. Recent research has uncovered the neural circuit mechanisms underlying high-order associative learning, which involves the transmission of state information between different circuits [13]. As shown in Fig. 1, in the cognitive systems of Drosophila, specific interneurons such as SMP108 play a critical role in transmitting already formed memory information from regions like α1 to other brain regions. Then, the formed memory acts as a reference, thus promoting the formation of high-order associative learning pathways. As a result, high-order associative learning exhibits a superior learning efficiency compared to low-order associative learning.

Based on these discoveries, we propose a high-order associative learning framework with a biologically realistic structure. Utilizing the synapse-like characteristics of memristors, this design integrates memristors as synaptic modules and incorporates a state calculation circuit to link low-order and high-order associative learning circuits. In Pavlov's classical conditioning experiments, the total power consumption of two memristors in the synaptic modules is below 11 $\mu$W, with a 230% improvement in learning efficiency compared to previous works. In image classification tasks, the features of images are captured in the states of a 20×20 memristor array within the low-order associative neural network. This state information is then transmitted to the high-order associative neural network, enabling accurate recognition and semantic labelling of test images with 100% accuracy. The scalability and versatility of this design, demonstrated across the above different application scales, highlights its potential to handle various cognitive tasks and enhance the efficiency of current memristor-based neuromorphic systems.

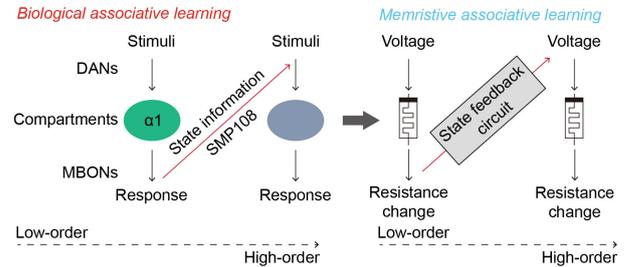

Fig. 1. High-order memristive associative learning circuits for efficient learning.

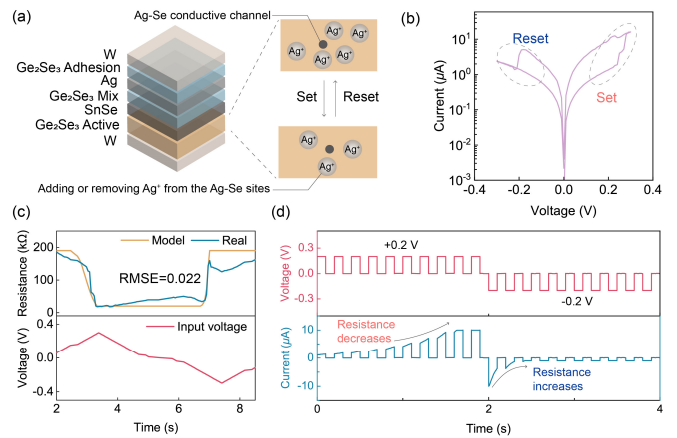

Fig. 2. (a) Structure and switching mechanisms of SDC memrsitors. (b) Hysteresis curve. (c) Comparison with physical SDC memrsitor and simulated model. (d) Pulse response measured from the simulated model.

## II. SELF-DIRECTED CHANNEL MEMRISTOR

To account for the general non-linear characteristics of memristive devices without focusing on specific structures and mechanisms, the self-directed channel (SDC) memristor (KNOWM Inc.) is purchased to build high-order associative learning circuits. The SDC memristor consists of multiple layers, as shown in Fig. 2(a) and its switching mechanisms can be found in [15]. As depicted in Fig. 2(b), its hysteresis curve exhibits an obvious 'set' process (from high resistance to low resistance) and 'reset' process (from low resistance to high resistance). Based on this switching behavior, the VTEAM model is used to simulate the SDC memristor for constructing the subsequent circuit. In this model, the switching behavior of SDC memristor is represented by an internal state variable, $w$, which is controlled by the external voltage. The derivative of the state variable $w$ is expressed as:

$$\frac{dw}{dt} = \begin{cases} k_{on}(\frac{v}{v_{on}}-1)^{\alpha_{on}} f_{on}(w), & 0 < v_{on} < v \\ 0, & v_{off} < v < v_{on} \\ k_{off}(\frac{v}{v_{off}}-1)^{\alpha_{off}} f_{off}(w), & v < v_{off} < 0 \end{cases} \quad (1)$$

where $v$ is the applied voltage stimuli, $v_{on}$ and $v_{off}$ are the positive and negative threshold voltages, respectively, and $k_{on}$, $k_{off}$, $\alpha_{on}$ and $\alpha_{off}$ are constants related to the switching characteristics of the memristor. The window functions $f_{on}(w)$ and $f_{off}(w)$ ensure that the state variable $w$ remains within defined boundaries. The relationship between memristance $R_M$ and state variable $w$ is expressed as the following (2) to capture the switching behaviour of the SDC memristor:

$$R_M(w) = R_{on}(\frac{R_{off}}{R_{on}})^{\frac{w-w_{on}}{w_{off}-w_{on}}} \quad (2)$$

where $R_{on}$ and $R_{off}$ represent the resistance in the low and high states, respectively, and $w_{on}$ and $w_{off}$ are the corresponding boundaries for the state variable $w$ in the ON and OFF states. Then, the Quasi-Newton method is applied to optimize the model parameters. During this process, the optimized Root Mean Square Error (RMSE) is defined as:

$$RMSE = \sqrt{(\frac{\sum((v_{model}-v_{real})^2)}{\sum(v_{real}^2)} + \frac{\sum((i_{model}-i_{real})^2)}{\sum(i_{real}^2)})/N} \quad (3)$$

where $v_{model}$ and $i_{model}$ represent the voltage and current calculated from the optimized VTEAM model, and $v_{real}$ and $i_{real}$ are the real voltage stimuli and current responses. The final fitting result achieves an RMSE of 0.022, as shown in Fig. 2(c). The optimized model parameters are listed in Table I. Based on this set of parameters, the simulated SDC memristor in the SPICE environment demonstrates clear synaptic behavior, as shown in Fig. 2(d).

TABLE I. PARAMETERS OF MEMRISTOR MODEL

| Parameter | Value | Parameter | Value |
|---|---|---|---|
| $R_{on}$ | 20 kΩ | $R_{off}$ | 190 kΩ |
| $\alpha_{on}$ | 1 | $\alpha_{off}$ | 1 |
| $k_{on}$ | 2.82 | $k_{off}$ | -18.33 |
| $v_{on}$ | 0.14 V | $v_{off}$ | -0.16 V |

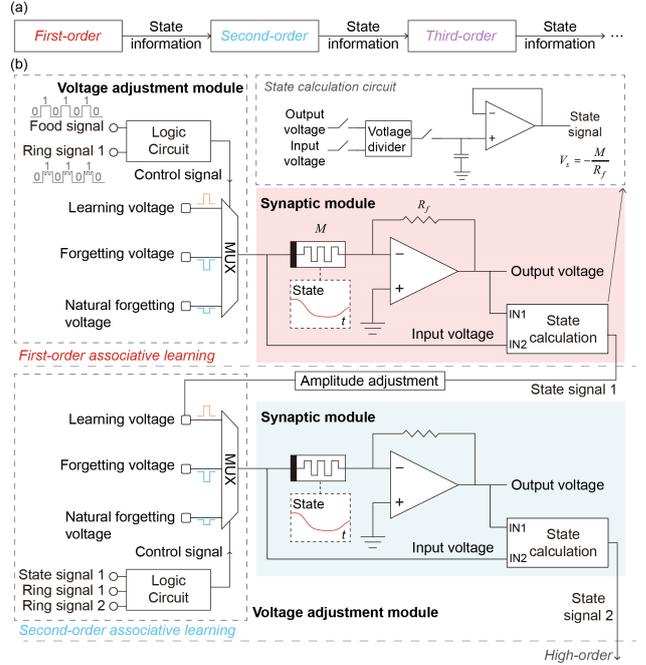

Fig. 3. Second-order associative learning circuit to demonstrate Pavlov's experiments.

## III. HIGH-ORDER ASSOCIATIVE LEARNING CIRCUIT

High-order associative learning refers to the process that links multiple stimuli through chains of associations [6]. For example, in Pavlov dog's experiments, first-order associative learning establishes a connection between a food signal and a ring signal, causing the dog to salivate at the sound of the ring after training. Second-order associative learning builds on this by associating the learned ring signal with a new signal. After sufficient training, this new signal can also elicit a response. Higher-order associations continue this process by linking newly learned signals with unknown stimuli. In our framework, the high-order associative learning framework is divided based on the order of association, as shown in Fig. 3(a). Each order consists of a voltage-adjustment module and a memristor-based synaptic module, with different orders connected through the state information of the memristor in the synaptic module. As shown in Fig. 3(b), in Pavlov's second-order associative learning, the first-order circuit establishes the relationship between the food signal and ring signal 1, and the learning process is reflected in the change of memristor state in the synaptic module. Then, this state information is delivered to the second-order associative learning circuit through the state calculation circuit, becoming a reference in adjusting the modulation voltage for the memristor in the second-order circuit which forms a connection between ring signal 1 and ring signal 2. Similar to other works, the food signal is encoded as square waves [16]-[18]. To differentiate signals, the ring signal is represented as square waves with zigzag variations in its high-level portion, reflecting different frequencies of the ring signal. Notably, these zigzag variations do not affect the subsequent logic detection, as the signal is still detected as a high-level voltage (logic 1).

### A. Voltage Adjustment Module

The voltage adjustment module consists of the logic circuit and analog multiplexer that analyze the sequential relationship between the input food signal and ring signal 1, and select the pre-defined modulation scheme. For instance, when both signals appear, the modulation scheme is set to the learning voltage. The relationship between modulation voltage and input signals is detailed in Table II.

TABLE II.  MODULATION SCHEMES CONTROLLED BY INPUT SIGNALS

| Food signal | Ring signal 1 | Modulation scheme |
|---|---|---|
| High-level | High-level | Learning voltage (0.35 V) |
| Low-level | High-level | Forgetting voltage (-0.175 V) |
| Low-level | Low-level | Natural forgetting voltage (-0.165 V) |

### B. Memristor-based Synaptic Module

The modulation voltage selected by the voltage adjustment module serves as the input for the memristor-based synaptic module. In this module, the positive terminal of the simulated SDC memristor is connected to the input voltage, while the negative terminal is connected to the inverting input of the operational amplifier (op-amp). This op-amp is configured with negative feedback using a feedback resistor ($R_f$). Due to the virtual ground created by the negative feedback, the input voltage ($V_i$) is applied precisely across the memristor, and the output voltage ($V_o$) can be expressed as:

$$V_o = -V_i \frac{R_f}{M} \quad (4)$$

where $M$ represents the resistance of the memristor in the synaptic module. When the food signal and ring signal 1 (to be associated) appear simultaneously, the positive learning voltage modulates $M$ to a low-resistance state, indicating the formation of an association between the food signal and ring signal 1. In addition to the learning voltage, the negative forgetting voltage and natural forgetting voltage modulate the memristor into a high-resistance state, as opposed to the learning voltage. To transmit the state information to the second-order associative learning circuit, a state calculation circuit involves a voltage divider designed to acquire the memristor state, as shown in Fig. 3(b). More details can be found in our work [19]. In this circuit, $V_i$ and $V_o$ serve as the divisor and dividend of the voltage divider circuit, and the inverted output voltage—representing the memristor's state signal in the first-order circuit ($S_1$)—can be expressed as:

$$S_1 = \frac{R_f}{M} \quad (5)$$

Subsequently, the amplitude of $S_1$ is adjusted using op-amps to serve as the learning voltage, and the adjustment needs to ensure that the learning voltage exceeds the positive voltage threshold while avoiding damaging the memristor during practical implementation. Then, the adjusted state signal serves as the learning voltage for the second-order associative learning circuit, mimicking that the learning efficiency of biological second-order associative memory is proportional to the strength of the first-order associative memory. In the second-order associative learning circuit, the circuit design is the same as the first-order except for the input signals, where the input signals are state signal 1 ($S_1$), ring signal 1 and ring signal 2. Notably, the threshold for determining whether state signal 1 is classified as high or low level is set at 0.1 V, corresponding to whether the memristor's resistance is below 50 kΩ. The specific relationship between the modulation scheme and the input signals is detailed as follows:

TABLE III.  MODULATION SCHEMES CONTROLLED BY INPUT SIGNALS

| State signal 1 | Ring signal 1 | Ring signal 2 | Modulation scheme |
|---|---|---|---|
| High-level | High-level | High-level | Learning voltage (adjusted $S_1$) |
| / | Low-level | High-level | Forgetting voltage (-0.19 V) |
| / | Low-level | Low-level | Natural forgetting voltage (-0.18 V) |

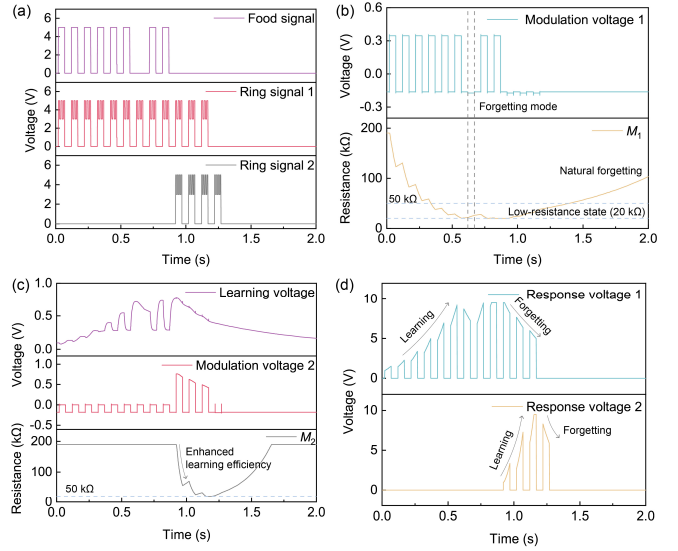

Fig. 4. (a) Input signals. (b) Modulation voltage for $M_1$ in the first-order associative learning circuit and the corresponding change in $M_1$'s resistance. (c) Learning voltage adjusted from $S_1$, modulation voltage for $M_2$, and the corresponding change in $M_2$'s resistance. (d) Response voltage.

## IV. RESULTS AND DISCUSSION

Based on the above circuit, the second-order associative learning in Pavlov's experiments has been demonstrated. As shown in Fig. 4(a), when the food signal and ring signal 1 appear simultaneously, the modulation scheme is set to the learning voltage. This causes the resistance of the memristor in the first-order associative learning circuit, $M_1$, to gradually decrease, thereby increasing the response voltage, which is calculated as the original input signal amplified by the change in $M_1$'s resistance. During this process, from 0.62 s to 0.67 s, only the ring signal 1 appears, prompting the modulation scheme to switch to the forgetting mode, as shown in Fig. 4(b). Consequently, $M_1$'s resistance increases slightly, resulting in a decrease in response voltage 1. When ring signal 1 and ring signal 2 appear together from 0.92 s to 0.97 s, and with $M_1$ already below the 50 kΩ threshold, the modulation scheme for the memristor in the second-order associative learning circuit, $M_2$, switches to the learning mode. Thus, the learning voltage adjusted based on $S_1$ modulates $M_2$ into the low-resistance state, as depicted in Fig. 4(c). Compared to the fixed learning voltage for $M_1$, this adaptive learning voltage, which is positively related to the strength of the low-order association, enables $M_2$ to enter the low-resistance state more rapidly, indicating higher learning efficiency. Unlike the same modulation time in previous designs [20], [21], the time required for $M_1$ and $M_2$ to fully switch to their low-resistance state decreases from 0.30 s to 0.13 s, representing a 230% increase in learning efficiency.

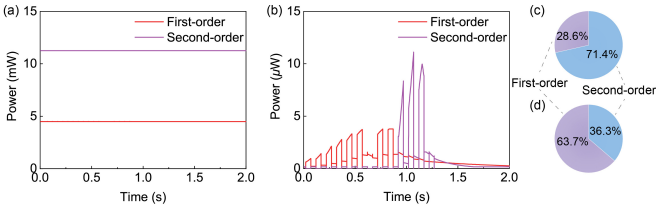

Fig. 5. (a) Power consumption of op-amps connected to memristors in the synaptic modules. (b) Power consumption of memristors. (c) Power comparison between op-amps in the first-order and second-order associative circuits. (d) Power comparison between memristors in the first-order and second-order associative circuits.

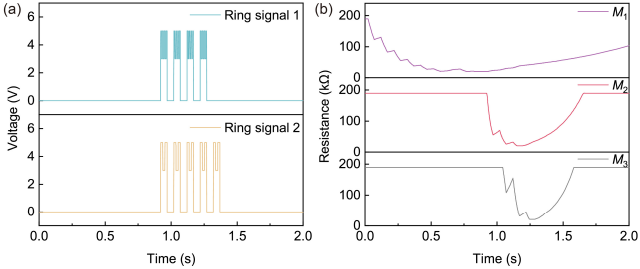

Fig. 6. (a) Ring signal 1 and additional input signal, ring signal 2, for the third-order associative learning circuit. (b) Resistance changes of the memristors in the synaptic modules.

This improvement is also reflected in the rapid increase of response voltage 2, as shown in Fig. 4(d). Upon removing external stimuli, $M_2$ quickly resets, replicating the transient nature of high-order associative learning in biological cognition systems. Regarding power consumption, in the synaptic modules, the op-amps connected to $M_1$ and $M_2$ consume 4.5 mW and 11.3 mW, respectively, as shown in Fig. 5(a). Meanwhile, the maximum power consumption of the memristors remains below 11 $\mu$W, as shown in Fig. 5(b) to Fig.5 (d). Following the proposed framework, the second-order associative learning circuit can be expanded to support higher-order associative learning (Fig. 6). By conveying the state signal of $M_2$ and integrating it into a third-second associative learning circuit, the memristor in the third-order circuit, $M_3$, exhibits even higher learning efficiency.

To demonstrate the broad applicability of our framework, we perform an image recognition task based on simulated memristive neural networks, as shown in Fig. 7(a). In this experiment, the state information transmitted from low-order associative learning enables the neural network to link high-level semantic information with input images, facilitating complex recognition tasks. Specifically, multiple groups of cat images from Kaggle datasets are firstly resized to 20×20 pixels. Subsequently, the input image is converted into a vector, and each pixel in the teacher image is compared with all values in the input vector. When the comparison condition is met, a counter is incremented. A modulation voltage positively related to this count is then applied to the corresponding location memristor in the memristor array, resulting in memristor resistance change, as shown in Fig. 7(b). After learning multiple groups of input images and teacher images, the normalized memristor array effectively represents a 'cat' through state information, as shown in Fig. 7(c). In second-order associative learning, this state information is compared with different categories of resized input images, allowing the system to construct the relationship between high-level features (semantic information) and the images. The similarity between the cat representation and the input image is formulated, where a lower value indicates a higher similarity [22]. A baseline similarity threshold is set, and by comparing the similarity to this threshold, a modulation voltage is generated to associate each input image with a label. As a result, ten input images are all accurately recognized and labeled as either 'cat' or 'non-cat'. Compared to other approaches (Table IV), our framework demonstrates superior efficiency and scalability while maintaining system compactness.

TABLE IV. COMPARISON WITH OTHER WORKS

|  | Our work | [23] | [18] | [17] | [24] | [25] |
|---|---|---|---|---|---|---|
| Low-order associative learning | √ | √ | √ | √ | √ | √ |
| High-order associative learning with superior efficiency | √ | × | × | × | × | × |
| Scalability | √ | × | × | × | × | × |
| Power consumption | 11 $\mu$W | 25 $\mu$W | NA | NA | NA | 0.1 $\mu$W |

## V. CONCLUSION

In this work, we propose a memristive high-order associative learning framework with biologically realistic details for the first time. By utilizing the synapse-like characteristics of memristors and bio-inspired structural design, this method achieves improvements in both learning efficiency (230% increase) and power consumption (with memristors in the synaptic modules consuming less than 11 $\mu$W) in Pavlov's experiments. Furthermore, the scalability and versatility of this design make it suitable for implementing high-order associative learning systems for cognitive tasks. To reproduce our results, circuits and codes needed are available at https://github.com/RTCartist/High-order-memristive-associative-learning.

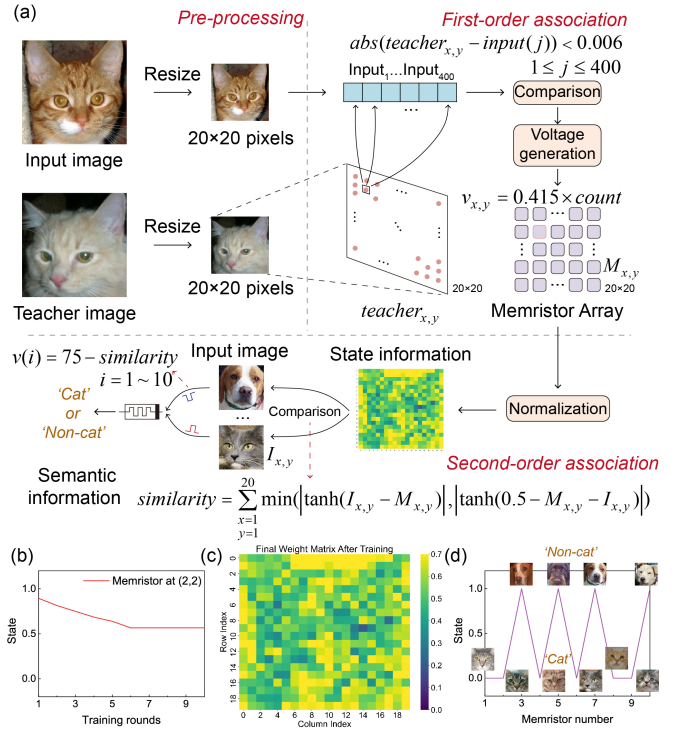

Fig. 7. (a) Structure of the associative neural network. (b) Resistance changes of the memristor at location (2,2) during training with ten groups of input and teacher images. State 0-1 corresponds to the normalized state following (2), where $w_{off}$ is 1, $w_{on}$ is 0. (c) State information. (d) Recognition results and semantic labelling, where the low-resistance state (0) represents 'cat' and the high-resistance state (1) represents 'non-cat.'

## REFERENCES

[1] S. Wang et al., "Memristor-Based Intelligent Human-Like Neural Computing," *Advanced Electronic Materials*, vol. 9, no. 1, p. 2200877, 2023.

[2] J. Tang et al., "Bridging Biological and Artificial Neural Networks with Emerging Neuromorphic Devices: Fundamentals, Progress, and Challenges," *Advanced Materials*, vol. 31, no. 49, p. 1902761, 2019.

[3] Z. Sun, S. Kvatinsky, X. Si, A. Mehonic, Y. Cai, and R. Huang, "A full spectrum of computing-in-memory technologies," *Nat Electron*, vol. 6, no. 11, pp. 823–835, Nov. 2023.

[4] W. Chen et al., "Essential Characteristics of Memristors for Neuromorphic Computing," *Advanced Electronic Materials*, vol. 9, no. 2, p. 2200833, 2023.

[5] S. Wang et al., "Memristor-based adaptive neuromorphic perception in unstructured environments," *Nat Commun*, vol. 15, no. 1, p. 4671, May 2024.

[6] J. M. Pearce and M. E. Bouton, "Theories of Associative Learning in Animals," *Annual Review of Psychology*, vol. 52, no. Volume 52, 2001, pp. 111–139, Feb. 2001.

[7] J. S. Biane et al., "Neural dynamics underlying associative learning in the dorsal and ventral hippocampus," *Nat Neurosci*, vol. 26, no. 5, pp. 798–809, May 2023.

[8] K. Nakahara et al., "Associative-memory representations emerge as shared spatial patterns of theta activity spanning the primate temporal cortex," *Nat Commun*, vol. 7, no. 1, p. 11827, Jun. 2016.

[9] H. An, Q. An, and Y. Yi, "Realizing Behavior Level Associative Memory Learning Through Three-Dimensional Memristor-Based Neuromorphic Circuits," *IEEE Transactions on Emerging Topics in Computational Intelligence*, vol. 5, no. 4, pp. 668–678, Aug. 2021.

[10] M. Yan et al., "Ferroelectric Synaptic Transistor Network for Associative Memory," *Advanced Electronic Materials*, vol. 7, no. 4, p. 2001276, 2021.

[11] Y. Zhang and Z. Zeng, "A Multi-functional Memristive Pavlov Associative Memory Circuit Based on Neural Mechanisms," *IEEE Transactions on Biomedical Circuits and Systems*, vol. 15, no. 5, pp. 978–993, Oct. 2021.

[12] Y. Pei, Z. Zhou, A. P. Chen, J. Chen, and X. Yan, "A carbon-based memristor design for associative learning activities and neuromorphic computing," *Nanoscale*, vol. 12, no. 25, pp. 13531–13539, Jul. 2020.

[13] D. Yamada et al., "Hierarchical architecture of dopaminergic circuits enables second-order conditioning in Drosophila," *eLife*, vol. 12, p. e79042, Jan. 2023.

[14] A. Handler et al., "Distinct Dopamine Receptor Pathways Underlie the Temporal Sensitivity of Associative Learning," *Cell*, vol. 178, no. 1, pp. 60-75.e19, Jun. 2019.

[15] K. A. Campbell, "Self-directed channel memristor for high temperature operation," *Microelectronics Journal*, vol. 59, pp. 10–14, Jan. 2017.

[16] J. Han, X. Cheng, G. Xie, J. Sun, G. Liu, and Z. Zhang, "Memristive Circuit Design of Associative Memory With Generalization and Differentiation," *IEEE Transactions on Nanotechnology*, vol. 23, pp. 35–44, 2024.

[17] J. Sun, Y. Wang, P. Liu, S. Wen, and Y. Wang, "Memristor-Based Neural Network Circuit With Multimode Generalization and Differentiation on Pavlov Associative Memory," *IEEE Transactions on Cybernetics*, vol. 53, no. 5, pp. 3351–3362, May 2023.

[18] Z. Wang and X. Wang, "A Novel Memristor-Based Circuit Implementation of Full-Function Pavlov Associative Memory Accorded With Biological Feature," *IEEE Transactions on Circuits and Systems I: Regular Papers*, vol. 65, no. 7, pp. 2210–2220, Jul. 2018.

[19] S. Wang et al., "Real-Time State Modulation and Acquisition Circuit in Neuromorphic Memristive Systems," *2024 IEEE Biomedical Circuits and Systems Conference (BioCAS)*, in press.

[20] B. Bannur, B. Yadav, and G. U. Kulkarni, "Second-Order Conditioning Emulated in an Artificial Synaptic Network," *ACS Appl. Electron. Mater.*, vol. 4, no. 4, pp. 1552–1557, Apr. 2022.

[21] G. Zhou et al., "Second-order associative memory circuit hardware implemented by the evolution from battery-like capacitance to resistive switching memory," *iScience*, vol. 25, no. 10, Oct. 2022.

[22] J. Y. S. Tan et al., "Monadic Pavlovian associative learning in a backpropagation-free photonic network," *Optica, OPTICA*, vol. 9, no. 7, pp. 792–802, Jul. 2022.

[23] Y. Yang, J. Mathew, R. S. Chakraborty, M. Ottavi, and D. K. Pradhan, "Low Cost Memristor Associative Memory Design for Full and Partial Matching Applications," *IEEE Transactions on Nanotechnology*, vol. 15, no. 3, pp. 527–538, May 2016.

[24] J. Sun, Y. Zhao, Y. Wang, and P. Liu, "Memristor-Based Affective Associative Memory Circuit With Emotional Transformation," *IEEE Transactions on Circuits and Systems II: Express Briefs*, vol. 71, no. 10, pp. 4601–4605, Oct. 2024.

[25] S. K. Vohra, S. A. Thomas, Shivdeep, M. Sakare, and D. M. Das, "Full CMOS Circuit for Brain-Inspired Associative Memory With On-Chip Trainable Memristive STDP Synapse," *IEEE Transactions on Very Large Scale Integration (VLSI) Systems*, vol. 31, no. 7, pp. 993–1003, Jul. 2023.